\documentclass[journal]{IEEEtran}
\IEEEoverridecommandlockouts
\usepackage{cite}
\usepackage{amsmath,amssymb,amsfonts}
\usepackage{algorithmic}
\usepackage{graphicx}
\usepackage{url}
\usepackage{hyperref}
\usepackage[ruled,vlined]{algorithm2e}
\ifCLASSOPTIONcompsoc
\usepackage[caption=false,font=normalsize,labelfon
t=sf,textfont=sf]{subfig}
\else
\usepackage[caption=false,font=footnotesize]{subfig}
\fi
\usepackage{caption}
\usepackage{tabularx}
\usepackage{textcomp}
\usepackage{xcolor}
\usepackage{array}
\usepackage{float}
 
\def\BibTeX{{\rm B\kern-.05em{\sc i\kern-.025em b}\kern-.08em
		T\kern-.1667em\lower.7ex\hbox{E}\kern-.125emX}}
\begin{document}

	\title{Smart Irrigation IoT Solution using Transfer Learning for Neural Networks}
	
	\author{\IEEEauthorblockN{Ali Risheh\textsuperscript{1}, Amirmohammad Jalili\textsuperscript{2} and Ehsan Nazerfard\textsuperscript{3}}\\
		\vspace*{0.5cm}
		\IEEEauthorblockA{\small{\textit{1 Department of Computer Engineering}, \textit{Amirkabir university}, Tehran, Iran, 
				ali.risheh@aut.ac.ir\\
		\textit{2 Department of Computer Engineering}, \textit{Amirkabir university}, Tehran, Iran, amirm.jalili@aut.ac.ir\\
				\textit{3 Department of Computer Engineering}, \textit{Amirkabir university}, Tehran, Iran, 
				nazerfard@aut.ac.ir}}}

	\IEEEabstract{
		In this paper we develop a reliable system for smart irrigation of greenhouses using artificial neural networks, and an IoT architecture. Our solution uses four sensors in different layers of soil to predict future moisture. Using a dataset we collected by running experiments on different soils, we show high performance of neural networks compared to existing alternative method of support vector regression. To reduce the processing power of neural network for the IoT edge devices, we propose using transfer learning. Transfer learning also speeds up training performance with small amount of training data, and allows integrating climate sensors to a pre-trained model, which are the other two challenges of smart irrigation of greenhouses. Our proposed IoT architecture shows a complete solution for smart irrigation. 
	}
	
	\IEEEkeyword{
		Automation Irrigation System, IoT, Machine Learning, Transfer Learning
	}
	
	\maketitle
	\section{Introduction}
	\subsection{Motivation}
	{\em Smart irrigation} is a major revolution in agriculture to replace costly and error prone human labor. In fact, smart irrigation is the most important part of smart agriculture due to limited water resources and importance of irrigation process on product quality. A particular sector of the agriculture industry which benefits most from smart irrigation is the greenhouse industry; this is because greenhouse crops are more sensitive to environment parameters, especially soil moisture, requiring real-time control and adjustment of irrigation faucets. Consequently, human operated irrigation of greenhouses is very costly and any human error results in large damage to the products.\footnote{In opposite to farmlands, human labors often do not live in proximity of the greenhouses.} While early solutions to reduce human cost are based on {\em remote irrigation} keeping human control in the loop, smart irrigation aims at providing a human-free automatic solution. However, a smart irrigation solution that addresses the practical challenges of the greenhouse industry is not currently available. 
	
	{\em Reliability} is the main practical challenge of any engineering solution to smart irrigation for greenhouses. In practice, the soil is divided into multiple layers, often $4$ or $5$ layers each with $10$ to $15$ cm thickness in our experiments. The moisture level of deepest layer should be constantly within $5$\% of its accepted level for a quality product, and any $20$\% deviation from the accepted level results in the crop death.
	
	{\em Internet of Things (IoT)} and {\em machine learning} are two technology advancements that enable smart irrigation. IoT enables system data collection via sensors and remote system control via actuators \cite{ghosh2016smart}, \cite{vaishali2017mobile}, \cite{saraf2017iot}. In the context of smart irrigation, an IoT system collects moisture level in all soil layers, and also other environmental variables such as temperature and LDR (light dependent resistor) to automatically control the irrigation faucets using IoT communication technologies. Machine learning enables accurate prediction of variables to control a system. For smart irrigation, machine learning allows prediction of soil moisture and appropriate control of the faucets. Artificial neural network models allow for high accurate prediction to satisfy greenhouse reliability constraints.
	There are three other practical challenges in applying machine learning in an IoT solution for greenhouses: limited data collection period (often a week or less), low processing power at IoT edge devices, and flexibility in adopting new environmental sensors. The first results in limited samples while training an artificial neural network from scratch requires high amount of training data. The last one requires flexibility of the machine learning (and the IoT solution) to add new sensors. {\em transfer learning} can address both of the above challenges. {\em Domain adaptation} in transfer learning allows adapting an existing model for other environments (different soil type, different type of greenhouse, etc) to a new environment using small training samples. It also reduces the training processing power needed. Moreover, {\em domain extension} in transfer learning allows adding new sensors by extending the structure of the existing artificial neural network.
	There is a small literature on machine learning and IoT for smart irrigation. \cite{capraro2008neural} uses recurrent neural network to predict required irrigation period. However, the model requires $\sim 1000$ hours as the IoT solution uses a single sensor on the top soil layer. \cite{murthy2019machine} uses a different approach by trying to replicate farmer's behavior by directly observing that. Of course, the farmer's behavior is not necessarily the best approach for irrigation and an IoT solution using sensory data can provide superior results.  \cite{regression} advocates adding temperature sensors to moisture sensors and runs a regression for prediction. We also add environmental sensors to our moisture data but use an artificial neural network for better prediction. IoT architecture for supporting smart irrigation is presented in 
	\cite{embedded} using Arduino and conditional states to control the faucets. However, the solution does not use full potential of embedded systems in state of the art IoT solutions which we will discuss.	
    \subsection{Contribution}
    Our goal is an automatic reliable and flexible irrigation for greenhouses discussed in the previous section. We provide a smart irrigation using an IoT based system plus a neural network model. We rely on moisture sensors in all soil levels ($4$ sensors) to achieve high accuracy in prediction, mean squared error (MSE) $< 0.05$. Our solution allows fast adaptation to new environments and flexibility in adding new environmental sensors using transfer learning. 
    
    To this end, we take the following steps (The rest of the paper is organized as following):
	\begin{itemize}
		\item We first collect data from two different soil types and show they have different characteristics. (section II.Data Description)
		\item Next, we train an artificial neural network for each soil separately that achieves our $0.05$ MSE target. We compare the results with a support vector regression (SVR) and show SVR requires higher training data to achieve the same MSE. (section III. Training)
		\item Next, we illustrate how to deploy transfer learning to use the neural network trained for soil one, for training a neural network for soil two using few of its samples. (section IV. Transfer Learning)
		\item Next, a transfer learning approach is presented to add new sensors. Particularly this approach allows the system to adapt itself without any disruption in its performance. (section V. Climate-Smart Agriculture)
		\item Finally, an IoT architecture will be introduced including end point sensors and actuators (faucet), a coordinator, a cloud server, and a communication protocol. The system is designed to be flexible for adding new sensors. (section VI. IoT Solution)
	\end{itemize}

	\section{DATA DESCRIPTION}\label{sec: data}
	In this section, datasets which were gathered from two different pots, denoted by Soil1 and Soil2, will be analyzed. Four soil moisture sensors are installed at a depth of $10$cm from each other in both soils. We denote these sensors by Moisture0 (in first layer), to Moisture 3. To better highlight the discrepancy, none of soils were irrigated during the time intervals. The data is stored once every 2-3 minutes. The total amount of data collected is 26,000 points. Our datasets are available online publicly, with updates of more samples at \cite{kaggleData}. Table \ref{dataset} is an example of our dataset.

\begin{table}[hbt]
		\scriptsize
		\caption{The Moisture0 in table is in the top layer of soil and the others increase in depth, respectively.}	
		\begin{tabularx}{0.47\textwidth} { 
  | >{\raggedright\arraybackslash}X 
  | >{\centering\arraybackslash}X | >{\centering\arraybackslash}X | >{\centering\arraybackslash}X | >{\centering\arraybackslash}X 
  | >{\raggedleft\arraybackslash}X | }
 \hline
year & month & day & hour & minute & second \\
 \hline
 2020 & 3 & 11 & 14 & 44 & 39\\
\hline
Moisture0 & Moisture1 & Moisture2 & Moisture3 & & \\
			\hline
			 0.59 & 0.63 & 0.51 & 0.45& & \\
			 \hline
\end{tabularx}
		\label{dataset}
	\end{table}
	
\begin{figure}[h]
        \centering
        \subfloat[Soil1 moisture]{\includegraphics[scale=0.6]{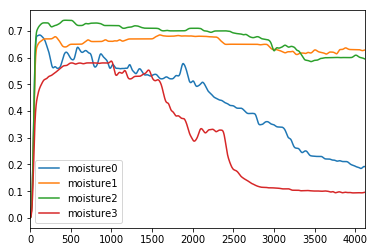}
      \label{fig:Soil1 moisture}}
      \hfill
    \subfloat[Soil2 moisture]{\includegraphics[scale=0.6]{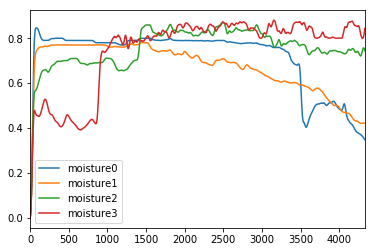}
      \label{fig:Soil2 moisture}}
      \caption{Two types of soil moisture comparison.}
         \label{moisturecomparison}
    \end{figure}

	In order to compare soils (Soil1 and Soil2) and show their differences, we first plot the moisture level for each soil (Figure ~\ref{moisturecomparison}).
	 The first layer of Soil1 loses water slowly, whereas in Soil2 the water remains longer in the first layer. We also see water passes second and third layers faster in Soil1, but in Soil2 the moisture slowly reaches the last layer. These show clearly that the rate of water absorption varies between the two pots. Moreover, the absorption rate across different layers of each soil are different (similar observations are made in \cite{soil-absorption}).
	
	\section{TRAINING}
	
	In this section, we train different models to predict the moisture in the deepest layer in the future from current moisture in all layers using two supervised learning methods: support vector regression and neural network. We denote the  period of collecting data with $T$ (2-3 minutes). The goal is to predict the moisture of the deepest layer at $3T$ ($6-9$ minutes ahead) to avoid overflow. So	inputs to the training model are $m_0^T,m_1^T,m_2^T,m_3^T$ and output of prediction is $m_3^{3T}$, where \(m_i^j\) is the moisture of layer $i$ in time $j$. All models are trained by sum of errors, in addition, MSE is represented as another metric to better show the convergence. MSE is multiplied by 10 in all graphs to better represent the trend.
	
	\subsection{Support Vector Regression}
	In this section we train a {\em Support Vector Regression (SVR)} model. Due to its simplicity, the model does not need high processing power and this system can be used on network edges specially in IoT communication technologies in which internet is not available. We will observe however the model requires large training samples for good performance. To address this shortcoming, we will study neural networks and transfer learning ideas in next sections.
	
	First we rewrite the problem statement in machine learning language with SVR model \cite{smola2004tutorial}. The prediction model is a linear function $f$ of the following form
	\begin{equation}
	f(x)=w \cdot x + b \hspace{0.5cm} with \hspace{0.5cm} w \in R^4,b \in R 
	\end{equation}
	The training problem is then the following optimization problem
	\begin{align}
	\min_{w\in \mathbb{R}^4} \sum_{i=1}^{N} (y_i- w \cdot {x_i}- b)
	\end{align}
 We first train the model using Adam optimizer
 with $4000$ samples in train set, $1000$ samples in test set and $20$ epochs. The model converges to local minimum after 20 epochs. The model and code are available in \cite{kaggleKernel}. 
	\begin{figure}[h]
		\centering
		\includegraphics[width=\linewidth]{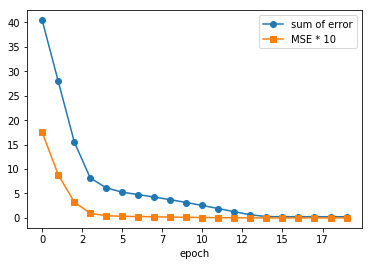}
		\centering
		\caption{Training error, both MSE and sum of errors, of SVR trained on $4000$ data point with Adam optimizer}
		\label{fig:SVR-fig}
	\end{figure}
	On convergence the model has the following test error which is in the acceptable range of $5\%$ error for us.
	\begin{center}
		Sum of error: 5.1476  \hspace{0.3\linewidth}         MSE: 0.0298.
	\end{center}
	While  with 4k data the model is achieving good enough performance for our application, the training model is very sensitive to number of samples. This is while in practice the number of samples often much less than 4k tested here. To see this sensitivity, we run the following experiment: we start from the pre-trained model for Soil1, and test it with 1k data from Soil2 (transfer learning). The test errors are:
	\begin{center}
		Sum of error: 20.3211   \hspace{0.3\linewidth}    MSE: 0.4255 .
	\end{center}
	This $40\%$ error in prediction is not acceptable in practice \cite{ideal-moisture}, therefore SVR can not be used. To address this issue, we first propose neural network for higher performance in prediction in section \ref{sec: ann}, second in Section \ref{sec: tl} we will discuss transfer learning which allows using a pre-trained model and adjust it with small samples to a new soil. 
	\subsection{Artificial Neural Network}\label{sec: ann}
\begin{figure}[h]
		\centering
		\includegraphics[width=\linewidth]{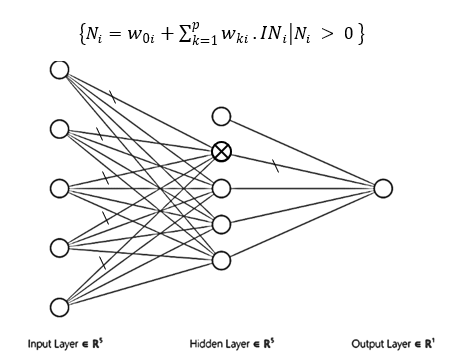}
		\caption{Schematic of neural network with dropped unit and disabled weights}
		\label{fig:nn_figure}
	\end{figure}

     {\em Artificial neural networks (ANNs)} are flexible computing frameworks and universal approximators that can be applied to a wide range of time series forecasting problems with a high degree of accuracy \cite{time-series-ANN}. Single hidden layer feed forward network is the most widely used model form for time series modeling and forecasting \cite{time-series-ANN-forecast}. From the result of SVR prediction we expect a single hidden layer neural network should suffice for our purpose. Figure \ref{fig:nn_figure} represents our neural network. The model is characterized by a network of three layers of simple processing units connected by acyclic links. The relationship between the output \(m_3^{3T}\) and the inputs \((m_0^T, m_1^T, m_2^T,\ m_3^T)\) has the following mathematical representation:
	\begin{equation*}
	m_3^{3T}= w_0+ \sum_{j=1}^{q}w_j.g\left(w_{0j}+\sum_{i=0}^{p}w_{ij}.m_{i}^T\right)
	\end{equation*}
	where, $w$ are model parameters often called connection weights; $p = 3$ is the number of input nodes (starts from 0) and $q = 4$ is the number of hidden nodes.
	Next we define the activation function which is appropriate for this model. Since moisture never becomes negative, we use the {\em Rectified Linear Unit (ReLU)} function.
	 \begin{equation*}
	     Relu(x) = max(0,x) 
	 \end{equation*}
	 We also define a dropout layer as a simple way to prevent neural networks from overfitting \cite{JMLR:v15:srivastava14a}. To this end, we choose a dropout rate to specifies the probability at which unit of the second layer is dropped. In Figure \ref{fig:nn_figure}, the crossed unit is dropped in a training case.
	
\begin{figure}[h]
		\centering
		\includegraphics[width=\linewidth]{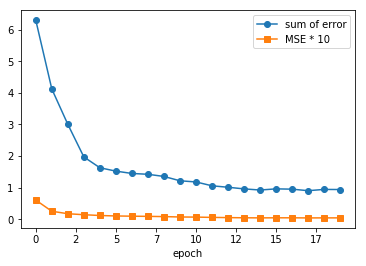}
		\caption{Training error, both MSE and sum of errors, of ANN trained on $4000$ data points with Adam optimizer}
		\label{fig:NN_result}
	\end{figure}

	Now we start training with same properties of training as previous part which is based on Adam optimizer using 4000 training data, 1000 test data with 20 epochs.
	Figure \ref{fig:NN_result} shows the training error of the neural network. Compared to SVR in Figure \ref{fig:SVR-fig}, neural network converges faster.
	
    Now we evaluate the test error for the neural network after 20 epochs:
	\begin{center}
		Sum of error: 0.2306    \hspace{0.3\linewidth}MSE: 8.4409e-05
	\end{center}
	The result has 3 order of magnitude higher accuracy compared to SVR and it does not do overfitting. 
	
	Neural networks are still sensitive to size of training data. For example with 480 samples (one day of sampling) the error results are the following which are above our reliability constraint.
	\begin{center}
		Sum of error: 8.2498    \hspace{0.3\linewidth}MSE: 0.1275
	\end{center}
	The second problem with neural networks is that they need higher processing power for training. Therefore, it is suitable for edge devices with high processing power. We address these issues and further improve the performance of neural networks with transfer learning in the next section.

	\section{TRANSFER LEARNING}\label{sec: tl}
	To address low performance of neural networks with small data and their high processing cost discussed in previous section, we use transfer learning that allows adopting the model from one soil to another soil while the two soils can have different distribution of underlying data. 

	Traditional data mining and machine learning algorithms make predictions on the future data using statistical models that are trained on previously collected labeled or unlabeled training data \cite{efficient_classification}, \cite{Classifier_Ensembles}, \cite{lazy_approach}. Semisupervised classification \cite{zhu2005semi}, \cite{text_classification}, \cite{Combining_Labeled}, \cite{Transductive_Inference} addresses the problem that the labeled data may be too few to build a good classifier, by making use of a large amount of unlabeled data and a small amount of labeled data. Variations of supervised and semisupervised learning for imperfect data sets have been studied; for example \cite{class-noise} have studied how to deal with the noisy class-label problems. Yang et al. considered cost sensitive learning \cite{test-cost_sensitive} when additional tests can be made to future samples. Nevertheless, these methods assume that the distributions of the labeled and unlabeled data are the same. This is not our case as shown in Section \ref{sec: data}. Transfer learning allows the domains, tasks, and distributions used in training and testing to be different \cite{transfer_leraning}. 
	
\begin{figure}[h]
	\centering
	\includegraphics[width=\linewidth]{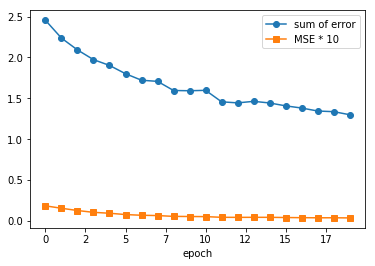}
	\caption{Train pre-trained model with 4000 new data of Soil2}
	\label{fig:ANN_model_1_2}
	\end{figure}

\begin{figure}[h]
        \centering
        \subfloat[New model]{\includegraphics[scale=0.3]{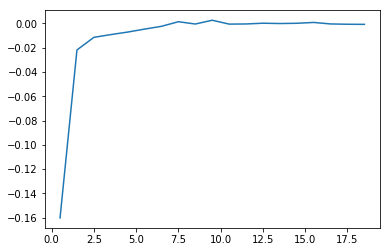}
      \label{fig:pre-trained_speed}}
        \hfil
    \subfloat[Pre-trained model]{\includegraphics[scale=0.3]{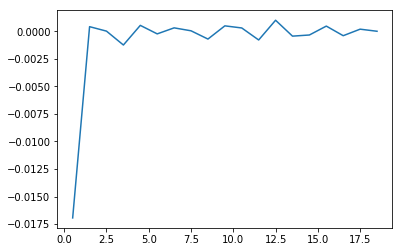}
      \label{fig:new_speed}}
        \caption{Pre-trained and new model comparison}
        \label{speed_comparison}
    \end{figure}

	We begin using Soil2 dataset.   Our pre-trained model (trained with 
	$4000$ training data of Soil1) with $1000$ test data of Soil2 gives the following error values:
	\begin{center}
		Sum of error: 2.2272    \hspace{0.3\linewidth}MSE: 0.0124
	\end{center}
	Now we run two experiments. First we train a new model from scratch with $4000$ training data of Soil2. 
	\begin{figure}[h]
	\centering	\includegraphics[width=\linewidth]{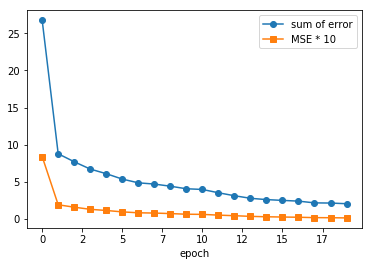}
	\caption{New model error result with Soil2 data by sum of error and MSE}	\label{fig:ANN_model_2_2}
	\end{figure}
	Figure \ref{fig:ANN_model_2_2} shows the results. Second, we transfer the model of Soil1 to Soil2 in the following way. We start from the model of Soil1 (trained in Section \ref{sec: ann}) and train with $4000$ samples from Soil2. The results are in Figure \ref{fig:ANN_model_1_2}.\\
	We can compare the two experiments. 
	First we compare their speeds of convergence which we define as difference in loss per epoch, depicted in Figure \ref{speed_comparison}.
    The pre-trained model has a faster convergence.	Second we compare the performance on initial points. Obviously the pre-trained model of experiment two has much better starting point performance. Third, we compare time to reach local minimum which is about 8 epochs for pre-trained model and 18 for experiment 1.	Comparing these results shows the transfer learning is superior in performance and achieves our practical reliability constraints ($MSE < 0.05$) at all times for the system.

    Next we run another experiment to assess the performance of transfer learning for our use case to compare the number of training data required by a new model. In each round, we add 50 samples to the number of training data of Soil2 and train a new model with previous conditions (epoch=20), then evaluate the 1000 test data of Soil2 and continue as long as the efficiency of the new model is the same as the pre-trained model with 4000 data of Soil1 ($until\hspace{0.2cm} sum \hspace{0.12cm} of \hspace{0.12cm} error = 2.2272$).
	\begin{figure}[h]
		\centering
		\includegraphics[width=\linewidth]{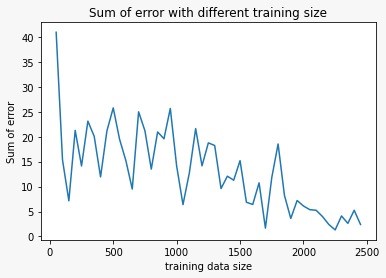}
		\caption{Accuracy by training data size from 50 to 2500}
		\label{fig:my_label}
	\end{figure}
	We have seen that a pre-trained model with 4,000 training data of Soil1, on 1000 test data of Soil2, had sum of error less than 3, this accuracy is possible with at least 2600 training data of Soil2 for a new model. This number of training data in our experiment indicates about 5 days of data collection to reach a new model to an acceptable accuracy. It should be noted that in our experiments, the data were collected every 2-3 minutes. In some cases, due to the lack of a power supply, this period may be increased to 30 minutes, in which case the need for a pre-trained model becomes more prominent.

    The use of transfer learning is very limited here. As mentioned, this method is also used to change the domain and the task. We have described a way to change the domain below. Our goal was to change the domain with the help of transfer learning methods, but due to our limitations in changing the model and data collection, it deprived us of this possibility.
	\section{CLIMATE-SMART AGRICULTURE}
	The {\em Climate-smart agriculture (CSA)} concept emerged in 2010 as a response to the imminent threat of climate change. In the original Food and Agriculture Organization (FAO) document (FAO, 2010), the deﬁnition of CSA is: “Agriculture that sustainably increases productivity, resilience (adaptation), reduces/removes greenhouse gases (GHG) and enhances achievement of national food security and development goals”. CSA has become known for the so-called “triple win”, i.e., working simultaneously to achieve its three objectives (or pillars): “adaptation, mitigation, and food security”. CSA is also sometimes presented as a mechanism to achieve synergies between the three pillars in a context-speciﬁc manner. CSA aims to contribute to sustainable landscapes and food systems as well as to resilience, ecosystem services, and value chains. Since the CSA concept did not arise from the academic community, its underlying concepts were not aligned with existing scientiﬁc debates, on e.g., sustainability, food security, resilience, or agroecology.
	This should not preclude CSA from being analyzed rigorously \cite{torquebiau2018identifying}.
	The field of CSA is very broad, we aim here to use climate variables to optimize irrigation, but this is in line with the concept of CSA because it aims at all three goals with the help of climate characteristics that will be discussed. As mentioned in the introduction, many variables (including temperature, light, etc) affect irrigation, our goal is to provide a way for each of these variables to contribute to the irrigation decision. It should be noted that the main model is based on the moisture of different layers. Conditions must be provided to which new variables can be added, and it is usual that not much data is available from the added variable(s). Suppose a variable such as air temperature is added to the model, in which case the amount of surface evaporation from the soil is affected by this variable, as well as the moisture of the different layers. The most convenient way to make this variable effective is to include it in the model. Changes in the structure of the neural network require a new training process, in which case model, predictions may be hampered by a lack of data from the new variable. The right solution is to train a new model and replace it at the right time, in the continuation of this discussion, a suitable solution is provided for this purpose. Mathematical expression of this method will be with the help of definitions in transfer learning concept.
	We assume we had a source domain \(D_s\) and a target domain \(D_t\).
	\begin{equation*}
	Ds=\left\{t_1,\ t_2,\ t_3,\ \ldots,\ t_n\right\} \hspace{0.1cm}and\hspace{0.1cm}Dt=\left\{X_0,\ t_1,\ t_2, \ldots,\ t_n\right\}
	\end{equation*}
	We have generated an artificial neural network (source model) for our $D_s$ and its data is complete and large. We make a copy of our source model with a new neuron in input layer and hidden layer (target model).
	\begin{figure}[h]
        \centering
        \subfloat[source model]{\includegraphics[scale=0.3]{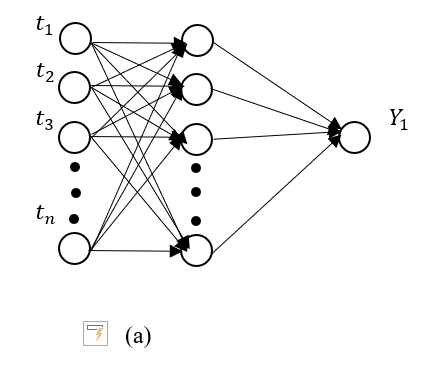}
        \label{fig:pre-trained_speed}}
        \hfil
        \subfloat[target model]{\includegraphics[scale=0.3]{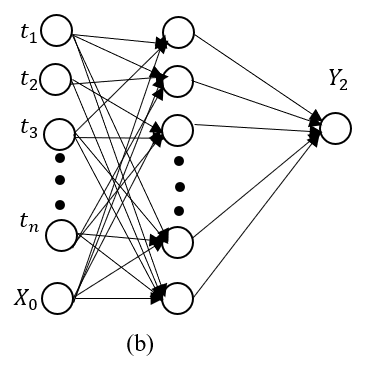}
        \label{fig:new_speed}}
        \caption{New model generated after new variable}
        \label{fig_derivatives}
    \end{figure}
    The method is clear from Figure ~\ref{fig_derivatives}, we just need to find a solution that can replace the new model (target model) at the right time with the old model (source model).
Now we start training the second model (new model) with new labeled data, but we use both models to predict final result based on linear regression with Y1 (source model prediction) and Y2 (target model prediction):
\begin{equation}
    {\alpha Y}_1+\ {\beta Y}_2=\ Y_{ans}
\end{equation}
\begin{figure*}[!t]
	    \centering
	    \includegraphics[height=7cm, width=13cm]{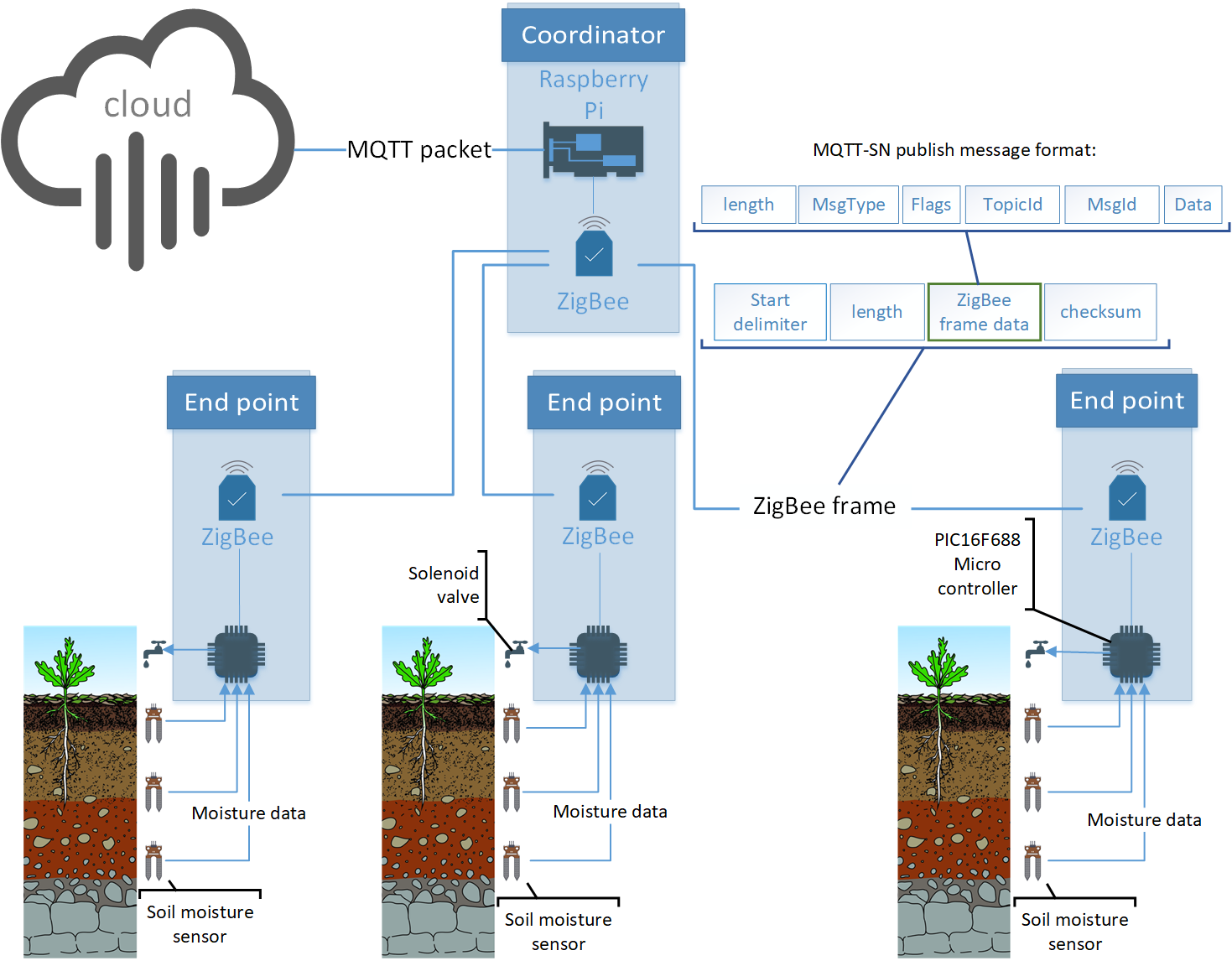}
	    \caption{An overview of the proposed system's architecture is depicted. The three major subsystems, namely endpoint, coordinator and cloud server are illustrated. Endpoints obtain moisture data from sensors and send them to the coordinator and carry out irrigation commands issued by the cloud server. A typical data frame structure exchanged between an endpoint and the coordinator is also shown.}
	    \label{architecturesolution}
	\end{figure*}
We initialize \(\alpha\) and \(\beta\) with $0.999$ and $0.001$ as we know our first model has been trained before. We will train our models base on \(Y_{ans}\)\ with gradient descend optimization and the linear regression separately.
\begin{algorithm}
\SetAlgoLined
Initialize \(\alpha=0.999\) , \(\beta=0.001\) \\
    while(True): \\
	\hspace{0.5cm}Y1 = Y (value from sensor) and train source model \\
	\hspace{0.5cm}Y2 = Y and train target model \\
    \hspace{0.5cm}\textbf{compute Y1 and Y2} \\
	\hspace{0.5cm}\(Y_{ans}\)=Y and optimize \({\alpha Y}_1+\ {\beta Y}_2=\ Y_{ans}\) \\
	\hspace{0.5cm}If \(\beta\) $> 0.9$ : \\
		\hspace{1cm}\textbf{Replace target model as the main model} \\
 		\hspace{1cm}\textbf{Break} \\
 \caption{Formula for add new variable:}
\end{algorithm}
In this section, we wanted to introduce a method so that we no longer need to provide a new system or model to add a new variable, this method helps each greenhouse to install a more accurate system according to its financial capacity. Our main method is to find the relationships between the soil layers, which will definitely be the safest way to predict soil moisture.
With this method, you can implement all the previous articles mentioned in the introduction \cite{murthy2019machine}, \cite{regression}, \cite{embedded}, and each person can add new option to the whole system only by paying for the desired sensor, without having to pay extra money.

	\section{IoT Solution}
	Smart Irrigation systems generally include three main technologies: evapotranspiration-based controllers, soil moisture sensor controllers, and rain-sensor-based controllers \cite{mccready2009water}. The proposed system in this paper is based on soil moisture sensor controllers. But, new variables like temperature data can be added to the system by simply connecting the required sensors.\\
	Various architectures for IoT-based agriculture systems using sensors to obtain soil data have been proposed in literature. In \cite{ghosh2016smart} different sensors were used to transmit soil data to a central server through serial communication and enable the user to monitor the data. In \cite{embedded} they used one moisture sensor per valve, Arduinos and a raspberry pi to make possible the automatic control of irrigation. Also, CherryPy server was used to connect the system to the web. In \cite{regression} LoRa P2P networks were used to automate the irrigation of crops. Soil data was collected using various types of sensors and then processed and sent to a server using a LoRa gateway.
	The smart irrigation system architecture proposed in this paper is illustrated in Figure \ref{architecturesolution}. The system consists of three major subsystems:
	\begin{itemize}
	\item \textbf{End points} collect soil moisture data to be processed by the cloud server. They are also connected to actuators to carry out irrigation.
	\item \textbf{Cloud server} is the subsystem that is responsible for receiving sensory data and instigating irrigation activities executed by end points.
	\item \textbf{Coordinator} is the center of a star topology. It acts as a mediator for communications taken place between server and endpoints.
	\end{itemize}
	\subsection{End Point}
	Each end point consists of moisture sensors embedded in different depths of the soil that provide sensory data, and Solenoid valves as actuators. Only one set of sensors per homogeneous parcel, where the soil and crop is homogeneous and constant, is used\cite{capraro2008neural}. Doing so makes the system more cost-effective. At the end point's heart is a low-power and low-cost microcontroller, namely PIC16F688. The end point perceives the irrigation operation as a binary decision sent by the cloud server and actuates it through opening or closing of faucets. Utilizing the microcontroller’s enhanced timer/counter, each end point implements a safety mechanism limiting the maximum irrigation time. When an irrigation operation starts, a timer is started and a predefined timeout is set to limit the maximum allowable irrigation duration. When the timeout is exceeded, irrigation is stopped and an error message is sent to the server. This prevents over-irrigation, water runoff, and plant damage in case of server failure. End points send sensory data to the coordinator by means of ZigBee devices. 
	\subsection{Coordinator}
	Coordinator manages communications between end points and the cloud server. It is the central node of a star topology. Its main component is a raspberry pi responsible for communicating with the cloud server and the end points. It is also capable of running python programs and, therefore, can carry out the irrigation management itself in case communication link to server is broken e.g. internet connection is down. For this to be realized, the prediction model must be downloaded to the raspberry pi. The raspberry pi communicates with end point devices using a ZigBee device.
	\subsection{communication design}
\begin{figure}[!h]
        \centering
        \includegraphics[scale=0.4]{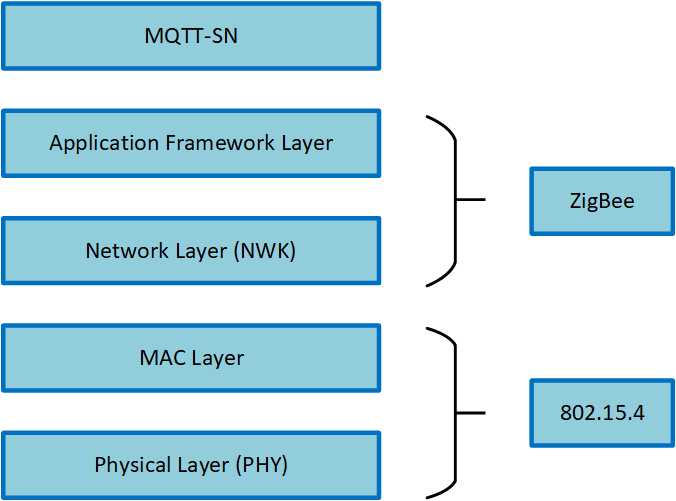}
        \caption {Network stack diagram. The physical and MAC layers are specified by IEEE 802.15.4 that ZigBee builds on. The network layer provides routing and establishes the star topology. Application framework layer is responsible for discovery and binding services. The highest-level layer is MQTT-SN protocol which enables the effective communication with the cloud server.}
        \label{protocolstack}
    \end{figure}

	One important aspect of wireless sensor networks is the method for communication between the nodes. Two prominent protocols are CoAP and MQTT. MQTT has been implemented for IoT-based irrigation systems and decent service availability has been observed \cite{murthy2019machine}. Using a common middleware, MQTT has been found to experience lower message delays than CoAP for lower values of packet loss \cite{thangavel2014performance}. since the underlying ZigBee network can support retransmissions up to four times in case data is lost, low data loss can be assumed, hence, MQTT will have lower delay for this particular architecture than CoAP. However, MQTT requires an underlying network, such as TCP/IP that provides an ordered connection capability.\\
    We decided to use MQTT-SN as the communication protocol for our network. MQTT-SN is a version of MQTT more adapted to wireless sensor networks, and is highly compatible with ZigBee networks \cite{stanford2013mqtt}. In our system, the end points are MQTT-SN clients, the coordinator is an MQTT-SN gateway and functions as translator between MQTT-SN and MQTT. Finally, the server has the role of MQTT broker. The system’s communication protocol stack is depicted in Figure \ref{protocolstack}.
    \section{Conclusion}
    We proposed a highly accurate IoT system for smart irrigation of greenhouses using artificial neural networks, using 4 soil sensors. We collected the data required for training the system, and proposed a transfer learning approach that can address the following issues: i) low number of training samples ii) low processing power at the edge devices iii) flexibility to adding new environmental sensors. Our transfer learning is based on using a pre-trained model to a different soil or set of sensory inputs for training a new model.  We showed the alternative state of the art SVR models do not satisfy the required accuracy, even with transfer learning. We are commercializing and implementing our techniques in real world under Sepantab start up company.
    There are several new directions. More experimentation using new collected field datasets can improve the techniques here. The approach can be extended to other environments such as farm irrigation. Also, the number of sensors can be adaptively chosen by the type of soil and accuracy level required to minimize costs.
\bibliographystyle{ieeetr}
\bibliography{references.bib}
\end{document}